\documentclass[a4paper, 10pt, conference]{ieeeconf}      
\usepackage{FG2021}
\usepackage{framed,multirow}

\usepackage{amssymb}
\usepackage{amsmath}
\usepackage{latexsym}
\usepackage{graphicx, graphics}

\usepackage{caption,subcaption}
\usepackage{url}
\usepackage[none]{hyphenat}
\usepackage{xcolor}
\usepackage{float}
\FGfinalcopy 

\IEEEoverridecommandlockouts                              
\def\FGPaperID{****} 

\title{\LARGE \bf
Weakly-supervised Joint Anomaly Detection and Classification
}

\author{\parbox{16cm}{\centering
    {\large Snehashis Majhi$^1$, Srijan Das$^3$, Francois Bremond$^2$, Ratnakar Dash$^1$ and Pankaj Kumar Sa$^1$}\\
    {\normalsize
    $^1$ Department of Computer Science and Engineering , National Institute of Technology Rourkela, India\\
    $^2$ INRIA Sophia Antipolis France \hspace{0.4in} $^3$Stony Brook University, USA}}
}

\begin{document}

\ifFGfinal
\thispagestyle{empty}
\pagestyle{empty}
\else
\author{Anonymous FG2021 submission\\ Paper ID \FGPaperID \\}
\pagestyle{plain}
\fi
\maketitle

\begin{abstract}
Anomaly activities such as \textit{robbery, explosion, accidents, etc.} need immediate actions for preventing loss of human life and property in real world surveillance systems. Although the recent automation in surveillance systems are capable of detecting the anomalies, but they still need human efforts for categorizing the anomalies and taking necessary preventive actions. This is due to the lack of methodology performing both anomaly detection and classification for real world scenarios. Thinking of a fully automatized surveillance system, which is capable of both detecting and classifying the anomalies that need immediate actions, a joint anomaly detection and classification method is a pressing need. The task of joint detection and classification of anomalies becomes challenging due to the unavailability of dense annotated videos pertaining to anomalous classes, which is a crucial factor for training modern deep architecture. Furthermore, doing it through manual human effort seems impossible. Thus, we propose a method that jointly handles the anomaly detection and classification in a single framework by adopting a weakly-supervised learning paradigm. In weakly-supervised learning instead of dense temporal annotations, only video-level labels are sufficient for learning. The proposed model is validated on a large-scale publicly available UCF-Crime dataset, achieving state-of-the-art results.
\end{abstract}

\section{Introduction}
Video surveillance is becoming more important in our daily life. Nowadays surveillance cameras are observed almost at every corner of cities. The primary goal of surveillance cameras is to increase security level and protect us from acts of violence, vandalism, terrorism, and theft. To ensure public safety, constantly monitoring these surveillance cameras on a 24/7 basis using manpower is a laborious and tedious task. Hence, development of automatic anomaly detection and classification methods for real-world surveillance videos is the utmost need of the hour. 

Many surveillance systems in the current scenarios have adopted computer vision based methods for anomaly detection. However, only detecting anomalies in real time is not sufficient for preventing the loss of human life and money. For this, timely categorizing the detected anomalies and enforcing right action to it can reduce the risk substantially. Many recent surveillance systems work on a semi-automatic fashion, where only anomaly detection is done by the system but categorization of anomalies require human efforts. This is due to the limited robustness of anomaly classification methods in real world scenarios. Moreover, recent approaches \cite{cvpr18,bmvc19} address the task of anomaly detection and classification independently. Hence the substantial improvement in anomaly detection methods leads to marginal progress in anomaly classification methods. In order to fully automatize the recent surveillance systems, development of a single model is required which can not only detect the anomaly correctly but also can help in classifying the anomalies for providing the right action in right time. Motivated by this, in this paper an end-to-end model is proposed that jointly learns detection and classification of anomaly instances in real world surveillance videos.

The task of joint anomaly detection and classification is challenging due to (i) unavailability of large corpus of annotated data for learning a wide variety of normal and anomaly patterns, and (ii) the traits of anomaly activities such as \textit{abuse, stealing, shoplifting, shooting, and robbery} differ in their temporal extent. The proposed model mitigates these challenges by adopting a weakly-supervised paradigm for the unavailability of temporally annotated video data and by designing a two-level attention mechanism to distinguish the anomaly activities which differs in their temporal extent. To learn from weak-supervision, Multiple Instance Learning (MIL)~\cite{mil} with ranking loss for anomaly detection has been adopted. By incorporating ranking loss, the model learns to maximize the margin of separation between anomaly and normal instances. In our model, a two-level attention mechanism is incorporated to focus on the video clips containing the anomalies for learning discriminative representation for anomaly detection and classification. The first-level attention mechanism highlights the clips pertaining to anomalies for the task of detection. Moreover, a relation between the anomaly detection and classification branch is developed through another level of attention mechanism. This second-level of attention mechanism enables the model to learn salient video representation for classifying anomalies. The proposed model is validated on a publicly available dataset, namely \textit{UCF-crime} achieving state-of-the-art results. In addition, baseline results are evaluated for joint anomaly detection and classification tasks introduced in this paper.

\begin{figure*}[h]
	\center
	\includegraphics[width=\linewidth]{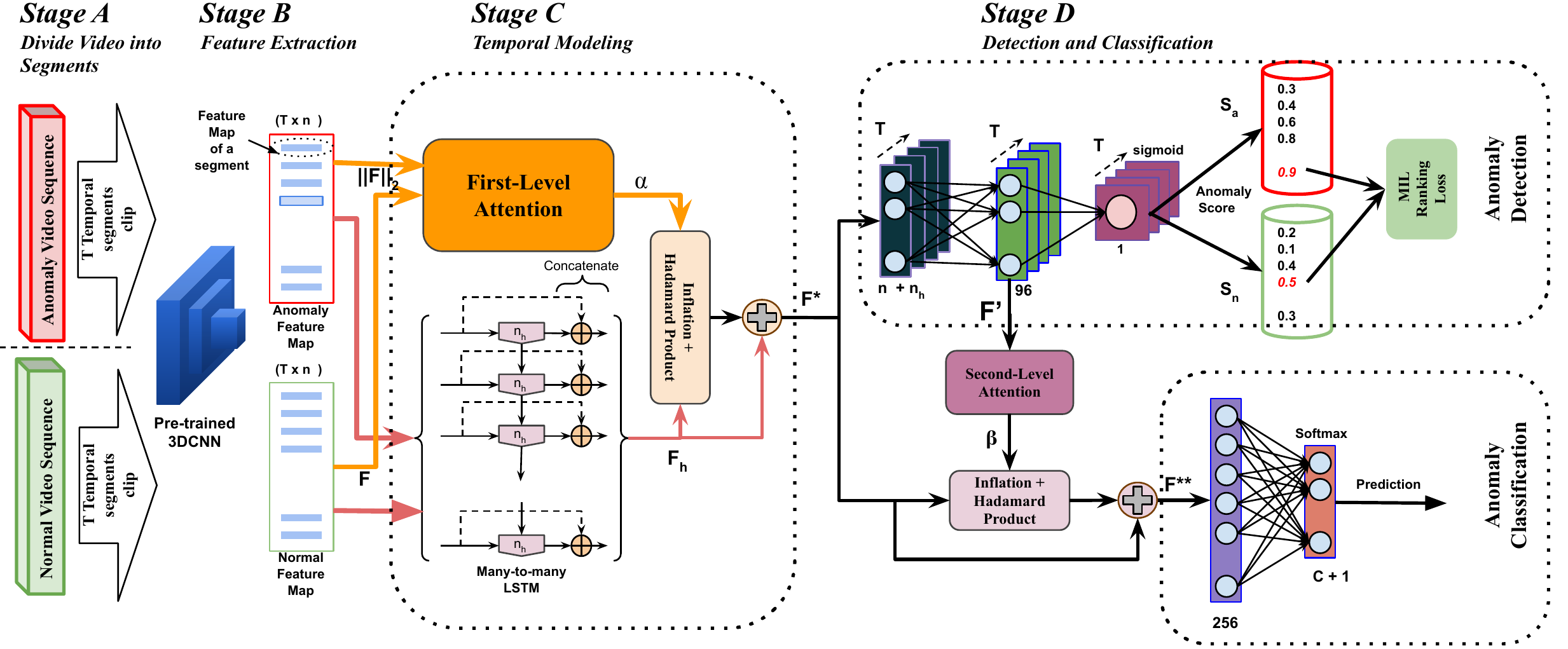}
    \caption{Our proposed Framework is composed of four stages: \textit{Stage A} divides anomaly and normal video sequences into $T$ fixed temporal segments. \textit{Stage B} extracts the feature from 3D CNN for each segment of the anomaly and normal videos generating respective feature maps. \textit{Stage C} consists of modeling the temporal evolution of the video segment features and a first-level temporal attention network. \textit{Stage D} is responsible for anomaly detection and classification. A second-level of temporal attention network is incorporated between the anomaly detection and classification to develop the inter-dependency among the tasks.}
    \label{framework}

\end{figure*}
In summary, the contributions of the paper include:
\begin{itemize}
    \item A novel model for joint anomaly detection and classification with weak supervision aiming to fully automatize recent surveillance systems. To the best of our knowledge, this is the first work to jointly deal anomaly detection and classification in a single model.
    \item A two-level attention mechanism at two different stages for discriminative video representation for anomaly detection as well as classification.
    \item Validation of the proposed model on a large-scale public dataset achieving the state-of-the-art results for classification while jointly performing the task of anomaly detection and classification.
\end{itemize}

\section{Related Work}

 Anomaly detection in surveillance videos is a challenging and active research topic in computer vision. Due to its massive application in video surveillance and crime scene investigation several research studies \cite{temporal_regularity,object_motion_pattern,HMM_spatio_temporal_volumes,crowded_scene,trajectories} have been reported in the last decade using hand-crafted features. Many earlier approaches \cite{OC_wacv2020,OC_gods_iccv2019,OC_PAMI2008,OC_cvpr2009,OC_pr2013,OC_kimcvpr2009,ST,Unsupervised_sparse_coding} address the problem of anomaly detection with one-class learning methods. These methods learn only from the normal patterns during training, defined for a specific scene. These approaches suffer from the fact that \textit{it is often difficult to learn feature representations for a wide diversity of normal patterns} and hence these methods are not suitable for general applications of anomaly detection.


To overcome the drawback of earlier methods, recent approaches \cite{cvpr18,icip2019,bmvc19,cvpr19} learn feature representation from normal and anomaly videos. This formulation of anomaly detection as binary class has been introduced with the dataset UCF-Crime \cite{cvpr18}. The success of deep learning models \cite{RC3d, SSN} in action detection motivates the researchers \cite{cvpr18,cvpr19} to make use of 3D convolutional networks (ConvNets) as visual backbone for segment wise feature extraction. By segments, we mean short video clips partitioned from an untrimmed video. Training 3D ConvNets like C3D, I3D pre-trained on huge datasets like Kinetics \cite{kinetics} and Youtube-8M~\cite{youtube8M} requires strong supervision for learning spatio-temporal patterns in a short video clip. Whereas obtaining temporal annotation for anomaly activities is a laborious task. Thus, \cite{cvpr18} addresses the task of anomaly detection under weak supervision which makes use of video-level annotation for untrimmed anomalous and normal videos. \cite{cvpr18} have proposed a MIL-based model to map video-segment based feature vector to an anomaly score. This mapping is learned through a ranking loss which optimizes the separation of the anomaly and normal segments in a video.
Inspired from previous studies, authors \cite{bmvc19} use the motion-aware features with MIL model to improve the anomaly detection. In addition, they also employ an attention block to incorporate temporal context while detecting anomalies.
Besides relying completely on the input features, the attention is applied only at the score level. Thus, the attention block does not modulate the feature maps leading to non-optimal detection. \cite{icip2019} model the temporal relation through Temporal Convolutional Network (TCN) \cite{TCN} and also proposed a novel complementary loss to maximize the margin of separation between inter and intra class instances. However, obtaining only long range temporal dependency is not sufficient for detecting anomalies in video. Thus for better understanding of temporal context, we propose a \textbf{first-level} of temporal attention mechanism to modulate the input feature to the MIL model. Another approach \cite{cvpr19} formulates the weakly-supervised problem as a supervised learning refining noise labels iteratively. But such methods are costly in terms of inference and is also data dependent, for instance relies too much on strong motion for detecting anomalies. 

Furthermore, no earlier approaches \cite{cvpr18,cvpr19,icip2019,bmvc19,icpr2020} address the task of anomaly detection and classification jointly. As a result, there is no significant progess in anomaly classification task. Mostly anomaly classification is treated as an independent problem in~\cite{cvpr18, bmvc19} where the videos are trained in an action recognition framework using visual backbones with video-level supervision. But these methods are not optimal as the visual backbones like TCNN~\cite{tcnn} and C3D~\cite{c3d} extract video-segment features and classify the whole untrimmed video into normal and anomalous samples through simple aggregation techniques. 
Thus, the classification of anomalies still remains a challenging task.

Inspirations for joint anomaly detection and classification is taken from action detection algorithms like~\cite{RC3d, SSN} but these algorithms are fully supervised with temporal intervals for each actions. In contrast, we proposed a joint detection and classification algorithm in weakly-supervised settings where the temporal intervals for an anomaly are not given. In order to design such an model, we incorporate a \textbf{second level} of attention mechanism between the anomaly detection and classification task, that builds an inter-dependencies among the tasks. 
Consequently, we propose a model with a two-level of attention mechanism which is jointly trained through global optimization for the task of anomaly detection and their classification.

\section{Anomaly detection and classification Model}
The proposed model shown in Fig. \ref{framework} for weakly-supervised joint anomaly detection and classification operates in two branches, where the first and second branch performs anomaly detection and classification respectively. From Fig. \ref{framework}, It may be observed that the proposed model consists of four stages and the stages are described below in detail.
\subsection{Divide Video into Segments}
In \textit{stage A} of Fig.~\ref{framework}, videos are divided into fixed number (say $T$) of temporal segments. Since the videos are untrimmed and no temporal annotations are provided, the video segments obtained from normal video sequences have all normal video clips whereas from anomaly video sequences have both anomaly and normal video clips. We assume that, there exists at least one anomaly segment among all video segments obtained from untrimmed anomaly video. 

\subsection{Feature Extraction}
In \textit{stage B}, spatio-temporal features are extracted from a 3D convolutional network for each video segment of anomaly and normal video sequences obtained in \textit{stage A}. Each video segment at time $t$ might have several clips of 64 frames and features are extracted for each such 64 frame clips. In order to have a fixed dimensional descriptor per video segment, a max pooling operation is performed across time. Thus, an $n$-dimensional feature vector $F_t$ is extracted for each video segment at time step $t$. Now, for a given video sequence has an output feature map of dimension ${T \times n}$, is used in the subsequent stages for the given task.

\subsection{Modeling Temporal Information}
Modeling of long-term temporal dependency is performed in \textit{stage C} through a Recurrent Neural Network (RNN) and a first-level temporal attention model.
In this stage, the extracted visual features are fed to a RNN, specifically LSTM in this case for modeling the temporal evolution of the anomalies in a video. The first-level temporal attention provides discriminative attention weights to the temporal segments those with probable occurrence of anomaly. Different from~\cite{bmvc19}, here modulation of features is done instead of decision scores for detecting anomalies. 

To obtain the temporal relationship among the segments in a video, a many-to-many LSTM $g()$ with $\theta_h$ parameters is used. The input to this LSTM are the feature vectors of the video $F \in \mathbb{R}^{T \times n}$ with $T$ temporal segments. The LSTM outputs a $n_h$-dimensional feature vector at each time step. Although temporal modeling of the video segments is achieved through LSTM but some visual information is lost in this process due to the sigmoid activated learning gates in the LSTM. So in order to retain the properties of the spatio-temporal feature map, we use a skip connection. Thus, the output at time step $t$ of the LSTM $F_h^t$ is computed as $F_h^t = h_t \oplus F_t$ where $h_t = g(h_{t-1},F_t;\theta_h)$ is the hidden state of LSTM. The symbol $\oplus$ represents concatenation operation between two variables. 

The LSTM has $n_h$ number of hidden neuron with \textit{tanh} squashing. So the output dimension of the LSTM followed by the concatenate operation for a given video is $T \times (n+n_h)$.

\textbf{ First-level Temporal Attention:} To highlight the temporal saliency to the LSTM feature vectors, a first-level attention weights is employed on them. Unlike \cite{bmvc19}, temporal attention weights are obtained from the same RGB feature map used for temporal feature learning.
\vspace{-.35cm}
\begin{figure}[h!]
	\includegraphics[width=\linewidth]{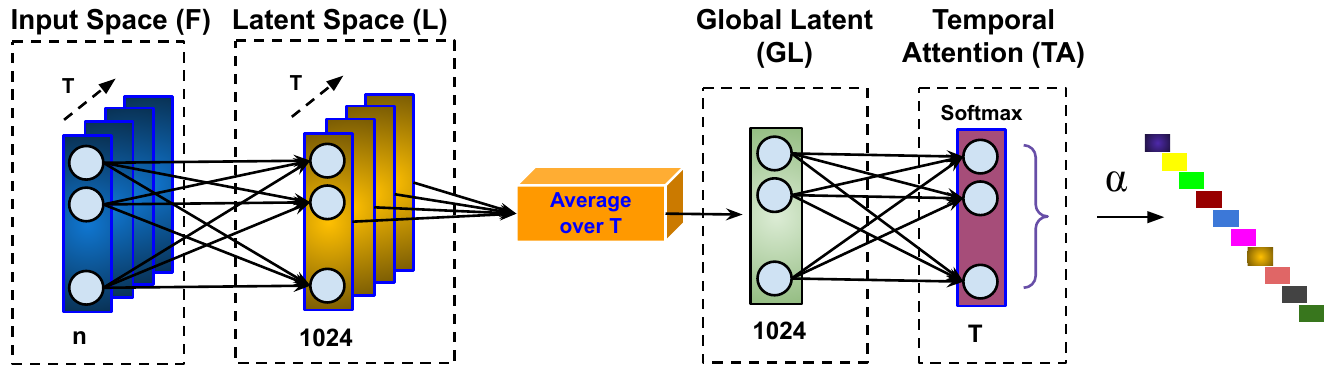}
    \caption{Proposed \textbf{First-level Temporal Attention Network} in details. This network learns attention weights $\alpha$ for every temporal segments from input spatio-temporal feature vectors.}
    \label{TA}
\vspace{-.5cm} 
\end{figure}

Figure \ref{TA} shows the detailed diagram of temporal attention model, where $F$ denotes the input feature vector from 3D CNN as mentioned earlier. The feature vector $F$ is first \textit{l}\textsubscript{2} normalized, then input to the first-level temporal attention network. In this attention network, instead of treating each segment of the video independently, a global representation of the video in a low dimensional latent space ($L$) is obtained using \texttt{time-distributed} perceptrons with weight vector $[W_{L,t}^1]_{t=1}^T$. 
This global representation of the video enables the model to consider the temporal context while learning appropriate weights for each temporal segment. 
 For input $F$ with $T$ temporal segments, the corresponding attention scores $s^1 = [s_t^1]_{t=1}^T$ are computed by

\begin{equation}
    \label{eqn_TL}
        \begin{aligned}
        s_t^1 = W_{GL}^1 \sum_{t=1}^{T}\frac{W_{L,t}^1 ||F_t||_2+b_{L,t}^1}{T}+b_{GL}^1\\
        \end{aligned}
\end{equation}
where $W_{GL,t}^1$, $W_{L,t}^1$ are the learn-able parameters with biases $b_{GL,t}^1$, $b_{L,t}^1$.
To obtain the temporal attention weights $\alpha_t$ at \textit{t\textsuperscript{th}} timestep, the scores (\textit{s\textsubscript{t}} \textit{$\forall$ t $\in$ T}) are normalized with \textit{softmax} activation.
 Thus, temporal attention weight at \textit{t\textsuperscript{th}} time step $\alpha_t$ is computed by $\alpha_t = \frac{\exp{(s_t^1)}}{\sum_{j=1}^T\exp{(s_j^1)}}$.

 \noindent Finally, the modulated feature vector $F^* = [F_t^*]_{t=1}^T$ is computed by adding the LSTM output feature vector with attention mask generated from Hadamard product of the attention weights and the LSTM output feature vector. The modulated feature vector for $t^{th}$ temporal segment is computed by $F_t^* = F_h^t +(\alpha_t F_h^t)$.

\subsection{Detection and Classification}
In \textit{stage D}, anomaly detection and classification is performed in two separate branches. The anomaly detection branch detect temporal segments containing anomaly patterns. Whereas, the anomaly classification branch classify the video to a specific anomaly class by its anomaly content. An inter-dependency between these two branches of the framework is also developed through a second-level attention mechanism. A detailed description of each branch is given below.

\subsubsection{Anomaly Detection}
The input to the \textit{anomaly detection branch} is the modulated feature vector $F^*$ from \textit{stage C}. In order to process each temporal segment of $F^*$ independently, \texttt{time-distributed} perceptrons are used. As shown in Figure \ref{framework}, this branch consists of 3 layer perceptrons. The final layer is \textit{sigmoid} activated to compute anomaly score for each temporal segment independently. 

\textbf{Second-level Temporal Attention:} This network builds a relationship between the detection and classification problem. Intuitively, the segments which produce high score in the anomaly detection branch must be an anomalous segment and such a segment should contribute more to recognize the anomaly than other segments. This functionality is performed by the second-level temporal attention network, which helps in highlighting the high scoring feature segments for the anomaly classification branch.
\vspace{-.5cm}
\begin{figure}[h!]
	\includegraphics[width=\linewidth]{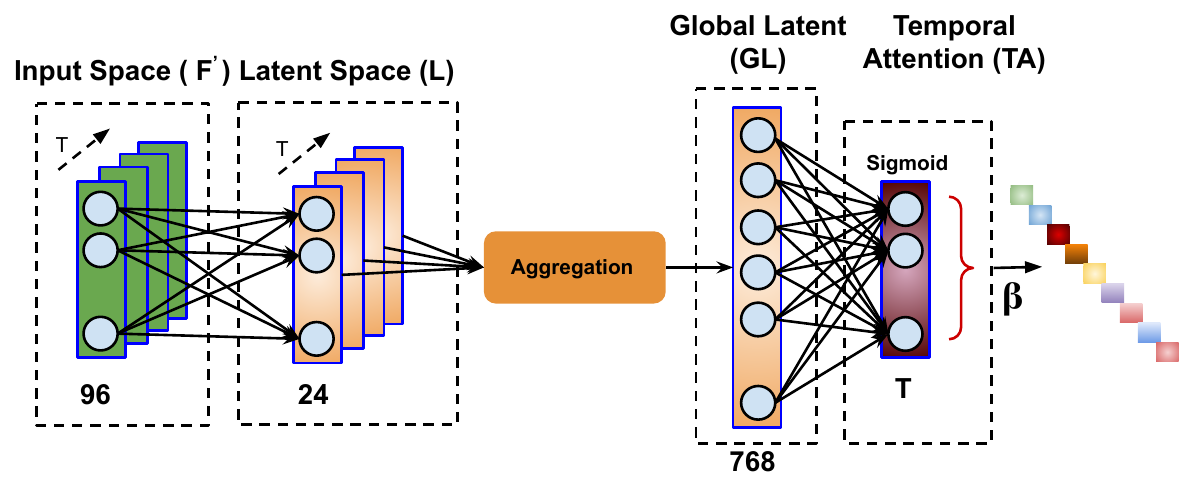}
    \caption{Proposed \textbf{Second-level Temporal Attention Network} for building inter-dependencies between anomaly detection and classification task. Note that Latent space $L$ and Global Latent space $GL$ here are different feature space from that in fig.~\ref{TA}.}
    \label{TSA}
\vspace{-.5cm} 
\end{figure}

Figure \ref{TSA} depicts our second-level temporal attention network. 
We aim at utilizing the features learning to detect anomalies for providing attention weights to the relevant temporal segments for the task of anomaly classification. Hence, a latent feature representation from the detection branch, the 2\textsuperscript{nd} layer of anomaly detection branch in our case is input to our second-level attention mechanism. We denote this latent features as $F'_{t} \in \mathbb{R}^{T\times n_L}$, where $n_L$ (\textit{=96, in our model}) is the number of neurons in the 2\textsuperscript{nd} layer of anomaly detection branch.

 For \textit{t\textsuperscript{th}} time step, the second-level attention score $s^2 = [s_t^2]_{t=1}^T$ is computed by a perceptron followed by aggregating the segment level features, and a final perceptron with neurons equal to the number of temporal segments $T$. Thus, the attention score is given by
\begin{equation}
    \label{eqn_TL}
        \begin{aligned}
        s_t^2 = W_{GL}^2 concat(W_L^2 F'_{t} + b_L^2) + b_{GL}^2
        \end{aligned}
\end{equation}
where $W_{GL}^2$,$W_{L}^2$ are trainable parameters with biases $b_{GL}^2$, $b_{L}^2$. The $concat()$ is a concatenate operation over time \textit{T} for aggregating features of \textit{T} time steps.

Finally, to obtain the second-level attention weight $\beta_t$ corresponding to \textit{t\textsuperscript{th}} temporal segment, we employ \textit{sigmoid} activation on the pre-computed attention score \textit{s\textsubscript{i}} given by $\beta_t = \frac{\exp{(s_t^2)}}{\exp{(s_t^2)+1}}$.
        
\subsubsection{Anomaly Classification}
The objective of \textit{anomaly classification} is to classify the video by the category of anomaly present in it. 
The input to the anomaly classification branch is the feature map $F^{**} \in \mathbb{R}^{T}\times (n + n_h)$ modulated by the second-level attention weight. The input to the anomaly classification branch at temporal segment $t$ is computed by $F^{**}_t = F^*_t + (\beta_t F^*_t)$.
 
 In order to classify $F^{**}$ to a specific anomaly class, a single feature representation is obtained by averaging the feature segments over time. This feature is used for classification using 2-layer perceptron. The final perceptron containing the neurons equal to the number of anomaly classes + 1 (for normal video) is \textit{softmax} activated to generate high prediction score for a specific class. 
 \vspace{-.25cm}
\subsection{Joint training all the networks}
Jointly training the two branches (\textit{i.e. anomaly detection and anomaly classification branch}) of the proposed model along with the two level attention network is a challenging task. In the proposed model, the anomaly detection branch follows a weakly-supervised learning and it is optimized based on a Multiple instance learning (MIL) ranking loss function proposed by \cite{cvpr18}, however the anomaly classification branch is optimized by categorical cross entropy loss with a video-level labels.

\paragraph{MIL Ranking Loss for detection}
For optimizing the anomaly detection branch, maximization of margin of separation among normal and anomaly instances is a crucial task. But identification of normal and anomaly instances without temporal annotations is difficult. Following the fact, \textit{an anomaly video contains at least one instance of anomaly and a normal video contains no anomaly}, \cite{cvpr18} aims at identifying normal and anomaly instances at score level. Since, prediction scores of anomaly and normal video instances are collected in \textit{S\textsubscript{a}} and \textit{S\textsubscript{n}}, a hinge-based ranking loss is enforced for maximization by treating the maximum score of \textit{S\textsubscript{a}} and \textit{S\textsubscript{n}} as anomaly and normal respectively. Along with the hinge based ranking loss, temporal smoothing and sparsity constraints are also proposed by \cite{cvpr18} given as follows:

\begin{equation}
    \label{eqn_NWD}
        \begin{aligned}
        Loss_D&(S_a,S_n) =\max{(0, 1 - \max_{i \in S_a}(S_a^i)+\max_{i \in S_n}(S_n^i)})\\
        &+ \lambda_1 \sum_i^{(N-1)}(S_a^i - S_a^{i+1})^2 + \lambda_2 \sum_i^N(S_a^i) 
        \end{aligned}
\end{equation}
where N = T $\times$ batchsize, $\lambda_1$ and $\lambda_2$ are the weighting factors of the temporal smoothing and sparsity constraints respectively. We employ the above ranking loss to dissociate the temporal segments in video into anomalies or normal video clip.

\paragraph{Cross Entropy Loss for classification}
In real-world, count of normal videos are far more than anomalous videos. So to handle such challenge of class imbalance while computing the classification loss, we employ a weighting factor equivalent to inverse class frequency in the loss.
We formulate the anomaly classification with the \textit{categorical cross-entropy} loss for \textit{C} classes of anomaly as:

\begin{equation}
    \label{eqn_reco}
        \begin{aligned}
         Loss_C = \sum_{i=1}^{C+1} (1-f_i)(y_i\log \hat{y_i})
        \end{aligned}
\end{equation}
where $f_i = \frac{\# samples in class_ i}{Total \# of samples}$. Note that the loss is computed for $C+1$ classes, one pertaining to normal videos. 

\paragraph{Regularization for attention}
We invoke a regularization term for the first-level attention weights that forces the model to pay attention to all the segments in the feature map. This is due to the fact that the attention weight is prone to ignore some segments in the temporal dimension although they contribute in detecting anomalous activities. Hence, we impose a penalty $\alpha_j\approx1$.
For the second-level attention, we employ another regularization term to regularize the learned attention weights $\beta$ with the $l_2$ norm to avoid their explosion.

\begin{equation}
        Loss_{att} = \sum_{j=1}^{T}(1-\alpha_{j})^2 + \sum_{j=1}^{T}||\beta_{j}||_{2}
\end{equation}
\paragraph{Optimization} In order to train the framework jointly, we need to optimize the total loss. The total loss can be represented as follows:

\begin{equation}
    \label{eqn_reco}
        \begin{aligned}
         Loss_{total} = \lambda_d Loss_D + (1-\lambda_d) Loss_C + \lambda_{att} Loss_{att}
         \end{aligned}
\end{equation}
where $\lambda_d$ and $\lambda_{att}$ are the weighting factors. 

\section{Experimental Analysis}

\textbf{Dataset description} - We have performed our experiments on UCF-Crime dataset \cite{cvpr18} containing real-world untrimmed surveillance videos. This is currently the largest anomaly detection dataset having a diversity in length of videos. The videos corresponding to a specific class can be of long ($approx$ 5 hrs) and short ($\leq$ 1 min) which makes the problem of detection more challenging. It contains 1900 videos out of which 950 videos are normal and rest are from 13 categories of anomalies. For training and testing 1610 and 290 videos are used respectively. As we are performing both anomaly detection and classification jointly, we follow the official anomaly detection train-test protocol provided by \cite{cvpr18}. Different from \cite{cvpr18,bmvc19} where the classification results have been provided on a separate uniform split of the dataset, we report the classification results on the same dataset split as for the detection task. Thus following this protocol, we provide new baseline for anomaly classification which can be used for follow-up comparisons.

\textbf{Evaluation Metric} - Since we are addressing two different tasks \textit{i.e. anomaly detection and classification}, two separate metrics have been used to evaluate our method. Following earlier works \cite{cvpr18}, we use frame-level Receiver Operating Characteristics (ROC) and its corresponding Area Under the Curve (AUC) for anomaly detection performance evaluation. For anomaly classification task, a mean-Average-Accuracy (mAA) is used where the average accuracy for each class is obtained and then mean of all average accuracies is reported to evaluate the performance.
\subsection{Implementation Details}
\paragraph{\bf Training}
For training, at first we extract the spatio-temporal features from I3D backbone~\cite{i3d} for each temporal segment. We set the temporal segments $T = 32$ in these experiments. 
For input to I3D, we take a center crop of dimension 224 $\times$ 224 from the full frame. The features are extracted from the \texttt{Global Average Pooling} layer of I3D which yields a feature of dimension $n = 7168$. 
For temporal modeling of the segments, the neurons in the LSTM is set to 1024 ($=n_h$). 
Then we train our end-to end trainable joint detection and classification model using Adam optimizer at a learning rate 0.0001 and with the loss weighting factors $\lambda_1=\lambda_2=8 \times 10^{-5}, \lambda_d$ = 0.9 and $\lambda_{att} = 0.1 \times 10^{-5}$. We also randomly select 30 positive and 30 negative videos as a mini-batch and compute the gradient using reverse mode automatic differentiation on computation graph using Theano. Then the loss is computed and back-propagated for the whole batch. 
\paragraph{\bf Testing}
During testing, we empirically found that center crop and the four corner crop is suitable for model testing and results in superior performance. This is to cover the fine detail of the anomaly, as in~\cite{cvpr19}.  
\subsection{Ablation study}

\begin{figure*}[h]
\scalebox{0.525}{
    \centering
    \begin{subfigure}[t]{0.3\textwidth}
        \centering
        \includegraphics[height=5cm,width=5cm]{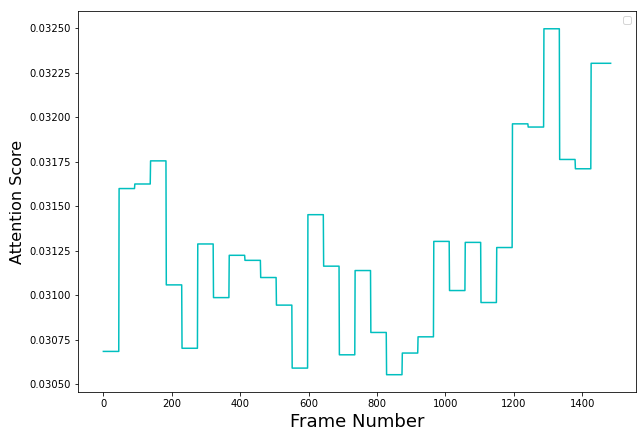} 
        \caption{}
    \end{subfigure}%
    ~ 
    \begin{subfigure}[t]{0.3\textwidth}
        \centering
        \includegraphics[height=5cm,width=5cm]{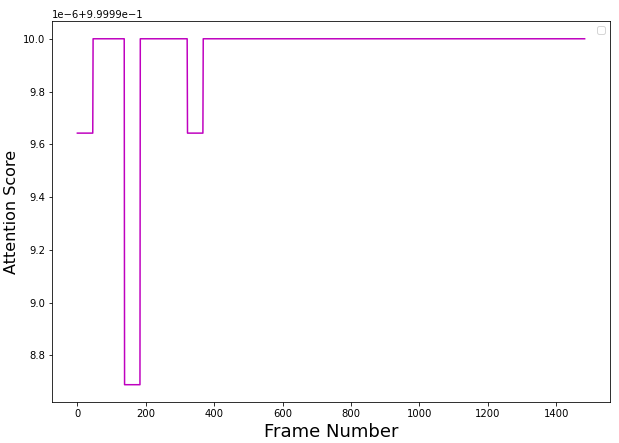} 
        \caption{}
    \end{subfigure}
     ~ 
    \begin{subfigure}[t]{0.3\textwidth}
        \centering
        \includegraphics[height=5cm,width=5cm]{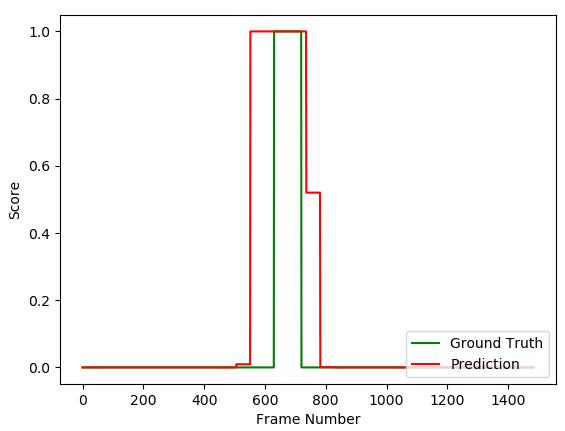} 
        \caption{}
    \end{subfigure}
    ~
    \begin{subfigure}[t]{0.3\textwidth}
        \centering
        \includegraphics[height=5cm,width=5cm]{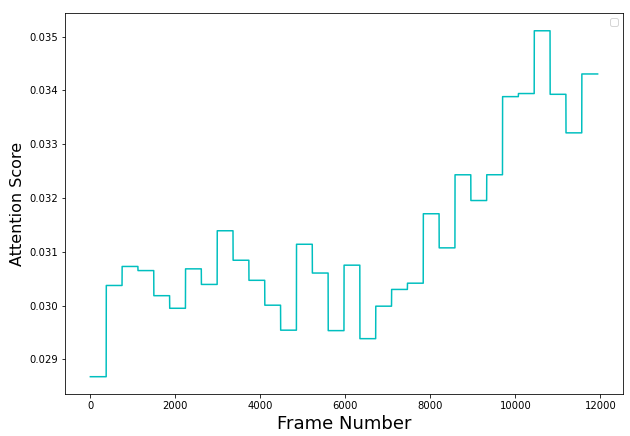} 
        \caption{}
    \end{subfigure}%
    ~ 
    \begin{subfigure}[t]{0.3\textwidth}
        \centering
        \includegraphics[height=5cm,width=5cm]{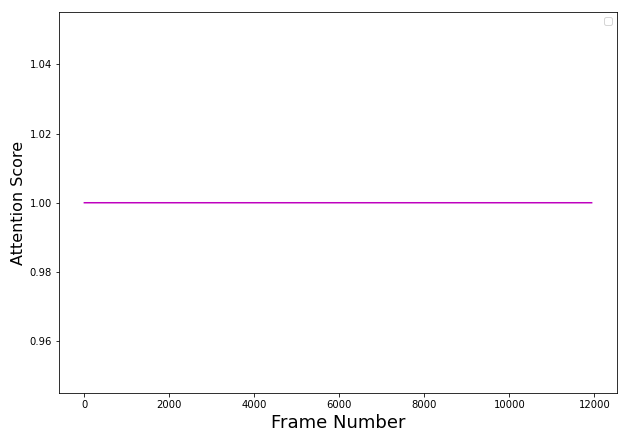} 
        \caption{}
    \end{subfigure}
     ~ 
    \begin{subfigure}[t]{0.3\textwidth}
        \centering
        \includegraphics[height=5cm,width=5cm]{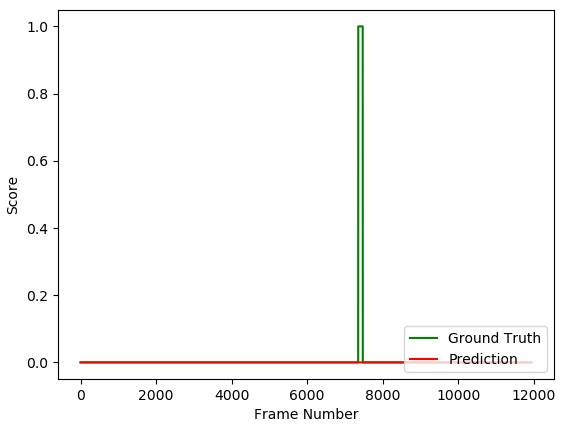} 
        \caption{}
    \end{subfigure}}
    \caption{Visualization of two-level of attention weights along with the prediction score. Figures (a)-(c) corresponds to the video "Shoplifting016" which is correctly detected and classified by the proposed model and Figures (d)-(f) corresponds to the video "Shoplifting034" which is miss detected and miss classified by the proposed model. Figure (a) and (d) shows the 1st level of attention weights,  Figure (b) and (e) shows the 2nd level of attention weights and Figure (c) and (f) shows the GT Vs. prediction scores of videos "Shoplifting016", "Shoplifting034" respectively.}\label{Abl_diagram}
    \vspace{-0.2in}
\end{figure*}

Table \ref{Abl_UCF} shows the importance of each component in our proposed model. The statistics show a significant improvement in detection and classification accuracy compared with I3D base model with or without two-level of attention. As a first baseline, we perform experiments with only I3D feature along with the classification and detection branch.
Prior to the two-level of attention, we incorporate LSTM for temporal modeling in untrimmed videos. It is visible from $l_2$ of Table \ref{Abl_UCF}, temporal encoding through LSTM improve detection accuracy by 1.52$\%$ (78.62$\longrightarrow$80.14) and classification accuracy by 5.36$\% $(14.66$\longrightarrow$20.02). 
Reversing the chronology of our proposed model, we first observe the influence of second-level attention weights in classification by adding the second-level attention network in the model to invoke an inter-dependence between the anomaly detection and classification branch, keeping the detection branch unchanged. It is evident from $l_3$ of Table \ref{Abl_UCF}, there is a significant improvement of 7.66$\%$ (20.02$\longrightarrow$27.68) in classification accuracy and a marginal improvement in detection accuracy. This is due to the ability of the second-level attention mechanism to assign high weights to all those video segments which contributes more to the classification of the video. 
Finally, we invoke the first-level attention network to verify the improvement in the anomaly detection task as well as substantiating the fact that \textit{boost in detection performance improves the classification task as well}.
Thus in $l_4$, we add the first-level attention network for highlighting the salient video segments containing possible anomaly pattern in the videos. The final row of Table \ref{Abl_UCF} indicates that the first-level attention weights boosts the detection accuracy by 1.73$\%$ (80.39$\longrightarrow$82.12) and it also leads to significant improvement in classification accuracy of 6.42$\%$ (27.68$\longrightarrow$34.1).

\begin{table}[t]
\begin{center}
	\caption{Ablation study on UCF-Crime dataset}\label{Abl_UCF}
	\begin{tabular}{|c|c|c|c|}
		\hline
		\bf Method	& \bf Components	& \bf AUC(\%) & \bf mAA	\\\hline
	    $l_1$	& I3D    &	78.62	&14.66\\\hline
        $l_2$	& $l_1$	+ LSTM  &   80.14   &   20.02	\\\hline
       $l_3$	&  $l_2$	+ Second-level attention &   80.39   &   27.68 \\\hline
      \bf  $l_4$	& \bf $l_3$	+ First-level attention   &    \bf 82.12   &   \bf 34.1\\\hline
	\end{tabular}
	\end{center}
	\vspace{-.95cm}
\end{table}

For qualitative analysis, we visualize the two-level of attention weights along with the prediction scores as illustrated in Figure \ref{Abl_diagram}. Interestingly, it is found that the proposed model fails to detect the anomalies in "\textit{Shoplifting034}" and also miss-classifies the video. In order to investigate about the "\textit{Shoplifting}" class, we found another video in the test set "\textit{Shoplifting016}" which is correctly detected and classified. To find out the possible cause of miss detection of video "\textit{Shoplifting034}", we analyzed the video and found that a lady is stealing a black color handbag which is similar to the handbag that she is carrying. 

This indicates that the attention mechanism derived from the \textit{appearance-only} features fail in providing reasonable attention weights due to possible confusion among the similar appearance of the bags. This implies that our model still needs an spatial attention for better understanding of the scene.

\begin{table}[t]
\begin{center}
	\caption{Anomaly detection and classification results on UCF-Crime dataset}\label{AUC_compare}
	\begin{tabular}{|c|c|c|}
		\hline
		\bf Methods	&	\bf AUC(\%) & \bf mAA (\%)	\\\hline
		SVM Baseline    &50 & - 	\\\hline
        Hasan \textit{et al.}\cite{temporal_regularity}  &   50.6 & - 	\\\hline
        Lu \textit{et al.}\cite{abnormal150} &   65.51 & -   \\\hline
        Sultani \textit{et al.}\cite{cvpr18}    &    75.41 & 13.9   \\\hline
        Dubey \textit{et al.}\cite{icpr2020} & 76.67 & - \\\hline
        I3D baseline & 78.62 & 14.6 \\ \hline
        Zhang \textit{et al.}\cite{icip2019}     &   78.66 & -   \\\hline
        Zhu \textit{et al.}\cite{bmvc19}     &   79 & -   \\\hline
        Zhong \textit{et al.}\cite{cvpr19}     &   82.12  & -  \\\hline
        \bf Proposed Model                      & \bf 82.12 & \bf 34.1   \\\hline
	\end{tabular}
\end{center}
\vspace{-.75cm}
\end{table}

\subsection{Comparison with state-of-the-art}
\begin{figure}[!h]
	\begin{center}
	    \includegraphics[scale=0.55]{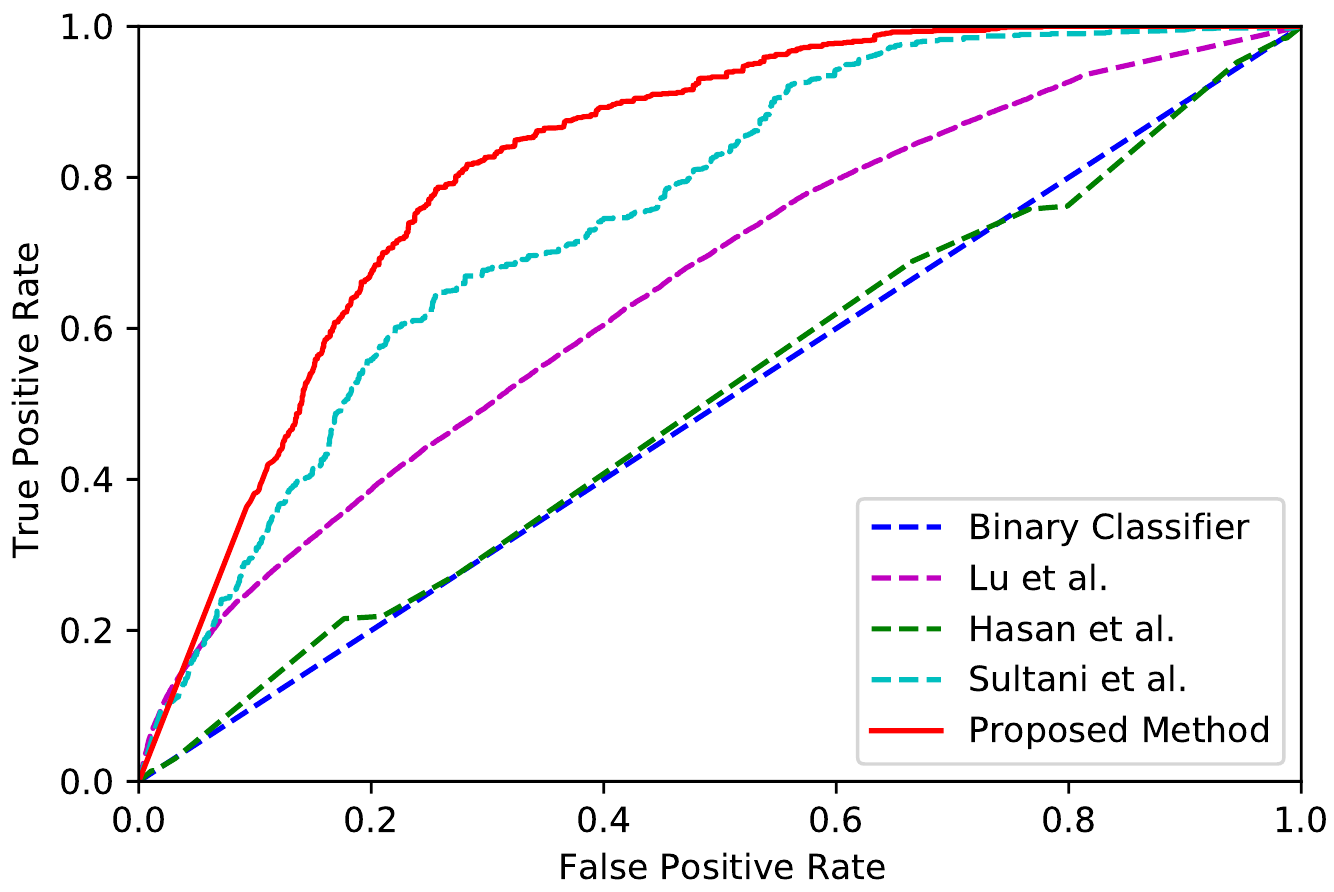}
	\end{center}
    \caption{State-of-the-art ROC curve comparison with the proposed method.}
    \label{ROC}
    \vspace{-.75cm}
\end{figure}

We have compared the recent state-of-the-art anomaly detection models on UCF-Crime dataset with our proposed model. We compare our model upon two indicators \textit{i.e.} ROC curves and AUC as shown in Figure \ref{ROC} and in Table \ref{AUC_compare}. 
As this is the first method to evaluate anomaly classification performance in detection split of UCF-crime dataset, we create baseline following the \cite{cvpr18} approach with the C3D \cite{c3d}, I3D \cite{i3d} backbone feature and K-NN classifier as shown in Table \ref{AUC_compare}. 
The authors in \cite{cvpr19} achieves similar results in anomaly detection to our model but they do not address anomaly classification problem in their method.
Earlier approaches lack in developing inter-dependencies between the anomaly detection and classification tasks, which resulted in anomaly detection performance not influencing the anomaly classification performance. In contrast, we formulated the two individual tasks jointly through an attention mechanism. Our attention mechanism significantly improves the classification performance by large margin while achieving state-of-the-art results for the task of detection as well. Following our approach, future developments in anomaly detection performance will lead to improvement in anomaly classification performance.

\section{Conclusion}

In this work, we proposed a weakly-supervised joint anomaly detection and classification model, driven by two-level of attention network. This model achieves state-of-the-art performance for the task of anomaly detection, and when detection and classification tasks are dealt jointly, it achieves significant improvement in anomaly classification performance. To achieve, temporal context modeling in weakly-supervised setting, we incorporate two-level of attention block in different stages of the model. From experimentation, it is evident the attention weights not only modulates the input features for anomaly detection but also helps in anomaly classification by developing an inter-dependency between anomaly detection and classification tasks. It is observed from the result analysis that our model still lags in detecting anomalies where real-world challenges like \textit{occlusion, background clutter, inadequate illumination} are present. \\

\vspace{0.1in}
\noindent\textbf{Acknowledgements.} 
This work has been supported by the French government, through the 3IA Côte d’Azur Investments in the Future project managed by the National Research Agency (ANR) with the reference number ANR-19-P3IA-0002. 
The authors are also grateful to the OPAL infrastructure from Université Côte d'Azur for providing resources and support.

{\small
\bibliographystyle{ieee}
\bibliography{my_bib}
}

\end{document}